\pdfoutput=1

\documentclass[a4paper,conference]{IEEEtran}
\usepackage{subcaption}
\usepackage{pdfpages}
\usepackage{paralist} % for \inparaenum
\usepackage{amsmath} % for \text
\usepackage[pagebackref=true,breaklinks=true,letterpaper=true,colorlinks,bookmarks=false]{hyperref}
\usepackage{algorithm,algorithmic} % for algorithm/algorithmic environment

\usepackage{etoolbox} 
\makeatletter
\patchcmd{\subsubsection}{\itshape}{\bfseries}{}{}
\makeatother

\definecolor{darkgreen}{RGB}{0, 64, 0}
\newif\ifhighlightchanges
\newcommand{\changemarker}[1]{%
  \ifhighlightchanges
    \textcolor{darkgreen}{#1}%
  \else
    #1%
  \fi
}

\ifCLASSINFOpdf
\else
\fi
\begin{document}
% \title{Automatic Visual Inspection of Cracks Using Controlled Illumination}
% \title{A New Crack Inspection Setup With a Comparative Study of Classifiers \& Illumination Conditions}
%\title{A New Crack Inspection Setup With a Comparative Study of Classifiers \& Height-Varying Illumination}
\title{A Versatile Crack Inspection Portable System based on Classifier Ensemble and Controlled Illumination}
%%V: I'd try to focus on the system rather than on the paper content

% \author{\IEEEauthorblockN{Milind G. Padalkar}
% \IEEEauthorblockA{School of Electrical and\\Computer Engineering\\
% Georgia Institute of Technology\\
% Atlanta, Georgia 30332--0250\\
% Email: http://www.michaelshell.org/contact.html}
% \and
% \IEEEauthorblockN{Homer Simpson}
% \IEEEauthorblockA{Twentieth Century Fox\\
% Email: homer@thesimpsons.com}
% \and
% \IEEEauthorblockN{James Kirk\\ and Montgomery Scott}
% \IEEEauthorblockA{Starfleet Academy\\
% San Francisco, California 96678--2391\\
% Telephone: (800) 555--1212\\
\IEEEoverridecommandlockouts % Required to unlock \thanks{} command
%%%%%% before acceptance %%%%
% \author{\IEEEauthorblockN{Milind G. Padalkar}
% \IEEEauthorblockA{Pattern Analysis and\\ Computer Vision,\\
% Istituto Italiano di\\ Tecnologia\\
% Genova, Italy\\
% milind.padalkar@iit.it}
% \and
% \IEEEauthorblockN{Carlos Beltr\'an-Gonz\'alez}
% \IEEEauthorblockA{Pattern Analysis and\\ Computer Vision,\\
% Istituto Italiano di\\ Tecnologia\\
% Genova, Italy\\
% carlos.beltran@iit.it}
% \and
% \IEEEauthorblockN{Matteo Bustreo}
% \IEEEauthorblockA{Pattern Analysis and\\ Computer Vision (PAVIS),\\
% Istituto Italiano di\\ Tecnologia\\
% Genova, Italy\\
% matteo.bustreo@iit.it}
% \and
% \IEEEauthorblockN{Alessio Del Bue}
% \IEEEauthorblockA{Visual Geometry\\ and Modelling,\\
% Istituto Italiano di\\ Tecnologia\\
% Genova, Italy\\
% alessio.delbue@iit.it}
% \and
% \IEEEauthorblockN{Vittorio Murino*\thanks{* V. Murino carried out this work when he was director of PAVIS at the Istituto Italiano di Tecnologia. He is currently also working for Huawei Technologies Co., Ltd., Ireland Research Center, Dublin, Ireland.}}
% \IEEEauthorblockA{Dipartimento di Informatica,\\
% %\IEEEauthorblockA{Pattern Analysis and\\ Computer Vision (PAVIS),\\
% University of Verona\\
% Verona, Italy\\
% %Istituto Italiano di Tecnologia\\
% %Genova, Italy\\
% vittorio.murino@univr.it}}
%%%%%%%
%%%%%% after acceptance %%%%
\newcommand*{\affmark}[1][]{\textsuperscript{#1}}
\author{
\IEEEauthorblockN{Milind G. Padalkar\affmark[1], Carlos Beltr\'an-Gonz\'alez\affmark[1], Matteo Bustreo\affmark[1,5], Alessio {Del Bue}\affmark[2]\affmark[*]\thanks{\affmark[*]The last authors A. {Del Bue} \& V. Murino contributed equally to the paper.} and Vittorio Murino\affmark[1,3,4]\affmark[*] \\
\{milind.padalkar, carlos.beltran, matteo.bustreo, alessio.delbue, vittorio.murino\}@iit.it
}
\IEEEauthorblockA{
\affmark[1] Pattern Analysis and Computer Vision (PAVIS), Istituto Italiano di Tecnologia, Genova, Italy \\
\affmark[2] Visual Geometry and Modelling (VGM), Istituto Italiano di Tecnologia, Genova, Italy \\
\affmark[3] Ireland Research Center, Huawei Technologies Co., Ltd., Dublin, Ireland \\
\affmark[4] Dipartimento di Informatica, University of Verona, Verona, Italy \\
\affmark[5] Dipartimento di Ingegneria Navale, Elettrica, Elettronica e delle Telecomunicazioni, University of Genova, Italy
}
}
%%%%%%%


%Telephone: (800) 555--1212, Fax: (888) 555--1212}


% \highlightchangestrue
% \input{sections/rebuttal}

\maketitle

\begin{abstract}
% The abstract goes here.
This paper presents a novel setup for automatic visual inspection of cracks in ceramic tile as well as studies the effect of various classifiers and height-varying illumination conditions for this task. The intuition behind this setup is that cracks can be better visualized under specific lighting conditions than others. Our setup, which is designed for field work with constraints in its maximum dimensions, can acquire images for crack detection with multiple lighting conditions using the illumination sources placed at multiple heights. Crack detection is then performed by classifying patches extracted from the acquired images in a sliding window fashion. We study the effect of lights placed at various heights by training classifiers both on customized as well as state-of-the-art architectures and evaluate their performance both at patch-level and image-level, demonstrating the effectiveness of our setup. 
\changemarker{More importantly, ours is the first study that demonstrates how height-varying illumination conditions can affect crack detection with the use of existing state-of-the-art classifiers. We provide an insight about the illumination conditions that can help in improving crack detection in a challenging real-world industrial environment.}
% \changemarker{More importantly, to our best knowledge ours is the first of its kind of study that addresses height-varying illumination conditions in combination with existing deep learning classifiers; in order to provide a state-of-the-art solution for an automated crack detection problem in a challenging real-world industrial environment.}
% To the best of our knowledge, this is the first study on height-varying illumination conditions for visual inspection of defects.
\end{abstract}


% \fi
\IEEEpeerreviewmaketitle

% \section{Introduction}
% % no \IEEEPARstart
% This demo file is intended to serve as a ``starter file''
% for IEEE conference papers produced under \LaTeX\ using
% IEEEtran.cls version 1.8b and later.
% % You must have at least 2 lines in the paragraph with the drop letter
% % (should never be an issue)
% I wish you the best of success.



% \subsection{Subsection Heading Here}
% Subsection text here.

% \subsubsection{Subsubsection Heading Here}
% Subsubsection text here.

% introduction
\section{Introduction}
\label{sec:introduction}

\begin{figure}
\centering
\begin{subfigure}[b]{\linewidth}
	\includegraphics[width=\linewidth]{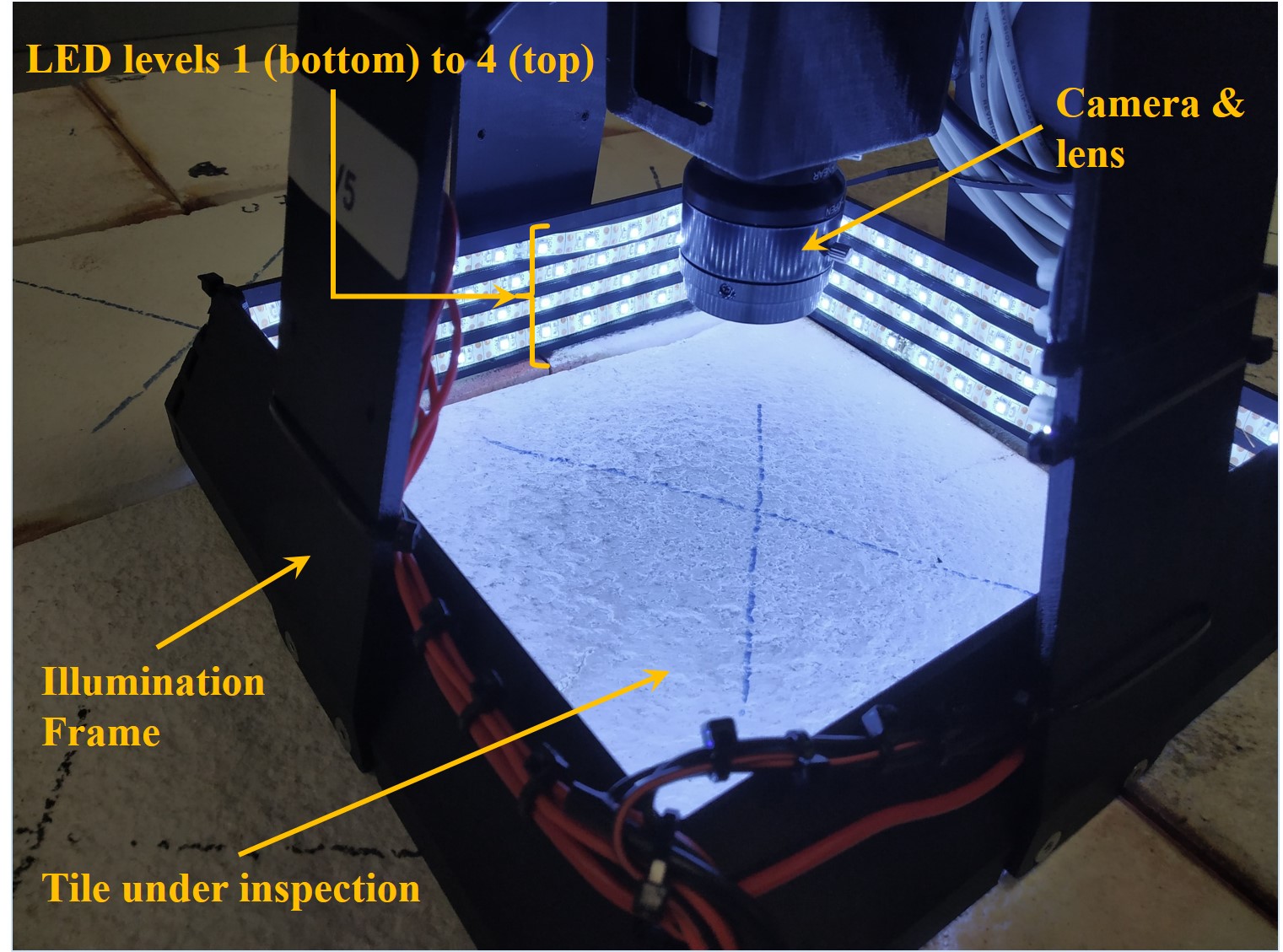}
	\caption{Illumination frame with the multiple LED levels and camera}\label{sfig:device_setup_lights}
\end{subfigure}
\begin{subfigure}[b]{.49\linewidth}
\includegraphics[width=\linewidth]{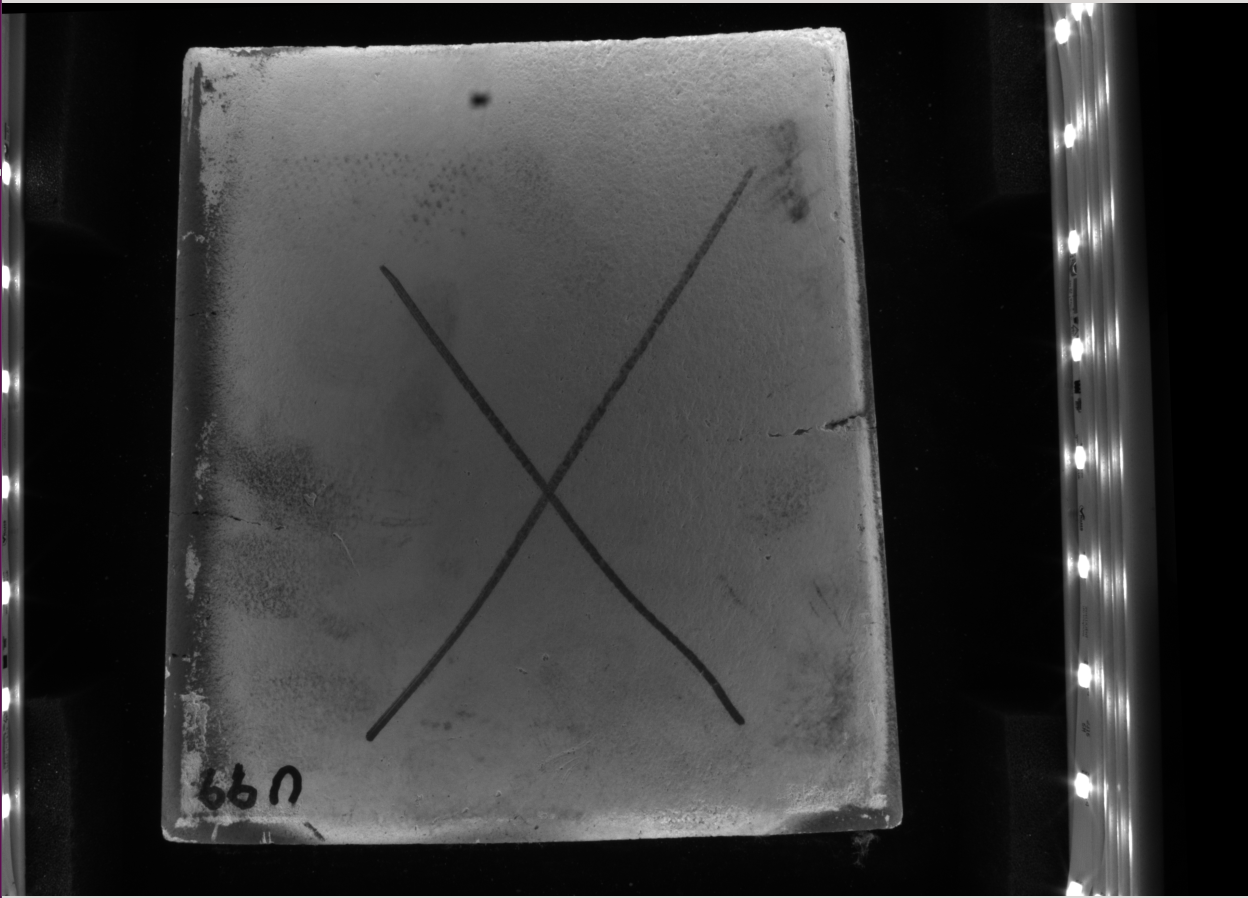}
   \caption{Acquired image}\label{sfig:example_input_image}
\end{subfigure}
%\begin{subfigure}[b]{.31\linewidth}
%\includegraphics[width=\linewidth]{./images/example_gt_cracks}
%    \caption{Annotated cracks}\label{sfig:example_gt_cracks}
%\end{subfigure}
\begin{subfigure}[b]{.49\linewidth}
\includegraphics[width=\linewidth]{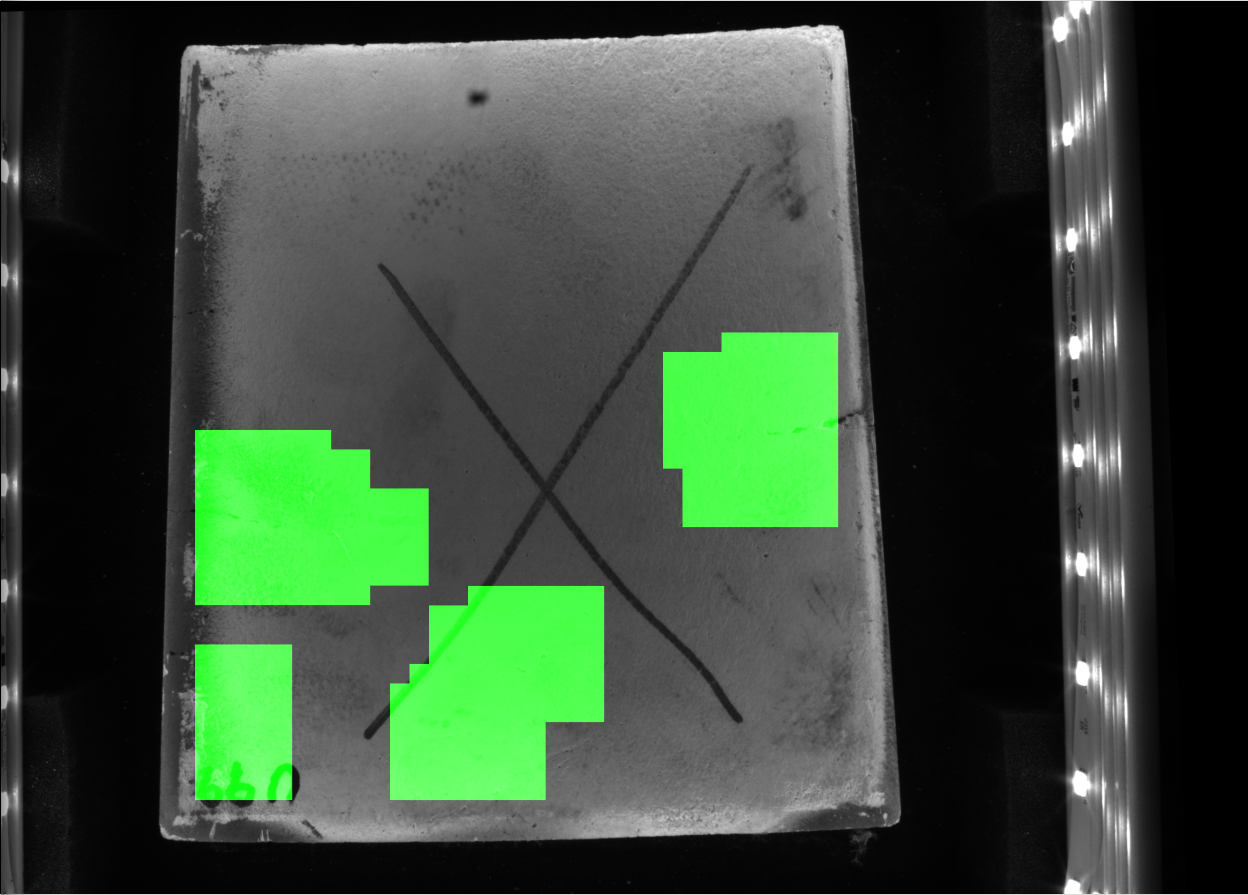}
    \caption{Automatically detected cracks}\label{sfig:example_detection}
\end{subfigure}
\caption{Crack detection in a ceramic tile using our setup with all the 4 LED levels switched on. (\subref{sfig:device_setup_lights}) Proposed setup; (\subref{sfig:example_input_image}) an acquired image using the setup and (\subref{sfig:example_detection}) automatically detected cracks in the acquired image.}
\label{fig:example}
\end{figure}

Visual inspection is an important step for ensuring quality of several industrial components. This is especially necessary in the case where a component requires maintenance and its failure could be catastrophic. In such a case, visual inspection is often performed by a human expert that is responsible for identifying the defective parts and suggesting their replacement, which they carry out based on their training and experience. Such a visual inspection process involves careful assessment of large number of parts, only few of which could present defects. Automating the visual inspection process can help in the assessment of large number of trivial cases, while the experts can be referred to only the non-trivial ones. Examples of such applications include identifying defects in casted steel \cite{Landstrom_DICTA_2013}, inspection of metallic components in nuclear power plants \cite{NB_CNN:Chen_TIE_2018}, detecting cracks in tiles that could be part of a leak-proof compartment, examining flaws in concrete structures \cite{Yang_IROS_2019},\ etc. Automating visual inspection in such applications involves acquiring digital images (or videos) of the areas to be inspected followed by defect detection using computer vision based algorithms.

Historically, the algorithms designed to detect defects like cracks followed a pipeline which typically involved contrast enhancement, edge linking and refinement \cite{Mohajeri_TRR_1991, Walker_TRR_1991, Feiguth_ICIP_99, Ammouche_CCC_2001}. The method proposed in \cite{Iyer_IVC_2005} also follows this approach wherein morphological features are used for edge extraction followed by refinement based on curvature. Similarly, the method in \cite{CrackTree:Zou_PRL_2012} uses pixel-neighbourhood statistics to identify crack pixels and tensor voting \cite{Medioni_RFIA_00} to connect them.

%\begin{figure}
%\centering
%\begin{subfigure}[b]{.49\linewidth}
%\includegraphics[width=\linewidth]{./images/example_input_image}
%   \caption{Acquired image}\label{sfig:example_input_image}
%\end{subfigure}
%\begin{subfigure}[b]{.49\linewidth}
%\includegraphics[width=\linewidth]{./images/example_gt_cracks}
%    \caption{Annotated cracks}\label{sfig:example_gt_cracks}
%\end{subfigure}
%\begin{subfigure}[b]{.49\linewidth}
%\includegraphics[width=\linewidth]{./images/example_gt_tile_area}
%    \caption{Annotated tile area}\label{sfig:example_gt_tile_area}
%\end{subfigure}
%\begin{subfigure}[b]{.49\linewidth}
%\includegraphics[width=\linewidth]{./images/example_detection}
%    \caption{Automatically detected cracks}\label{sfig:example_detection}
%\end{subfigure}
%\caption{Crack detection using our setup by switching on all the illumination sources. The automatically detected cracks in (\subref{sfig:example_detection}) are at the same locations as those shown in (\subref{sfig:example_gt_cracks}) which are annotated by an expert.}
%\label{fig:example}
%\end{figure}

Over the past decade, deep learning based methods have gained a lot of popularity given their performance on various vision tasks. In the deep learning framework, the crack detection problem is now generally posed as classification task, where a label (among crack/no-crack) needs to be predicted for every pixel (or patch) of the given image. Labeling of every pixel in the image is performed using encoder-decoder based architectures derived from the method proposed in \cite{SegNet:Badrinarayanan_TPAMI_2017}. On the other hand, architectures designed for image classification (inspired from AlexNet \cite{AlexNet:Krizhevsky_NIPS_2012}) can be used to classify the image patches.

The methods proposed in \cite{Yang_CVPRW_2018, Dong_ECCVW_2018, Yang_IROS_2019, Yang_TrITS_2019} follow the former approach of labeling every pixel in the image. These networks have a pyramidal form following the U-Net architecture \cite{UNet:Ronneberger_MICCAI_2015}. Additionally, some of them also have side outputs that act as edge priors \cite{HED:Xie_ICCV_2015} for crack detection. However, if the network architecture is not fully convolutional, then the input image is required to be down-scaled to match the respective network's input size. This can lead to loss of image resolution due to which details of fine cracks can be lost. On the contrary, the techniques proposed in \cite{NB_CNN:Chen_TIE_2018, Park_JCCE_2019} follow the latter approach. Here, the advantage is that only specific image areas can be inspected for the presence of cracks by selecting the appropriate patches using inexpensive pre-processing. Also, to match the network's input size, only the patches need to be resized as opposed to resizing the complete image as done in the pixel-labeling based approaches. Thus, the resolution of the inputs provided to the networks in the patch-classification based approaches can be higher in comparison to the pixel-labeling based approaches.

In this paper, we follow the latter approach and perform crack detection by classifying patches extracted from the image in a sliding window fashion.
%It may be noted that we do not propose a new architecture for crack detection but compare the crack detection results by using state-of-the-art classification architectures \cite{VGG16:Simonyan15, Xception:Chollet_2017_CVPR, ResNet50:He_2016_CVPR, DenseNet121:Haung_2017_CVPR, InceptionResNet-v2:Szegedy_2017_AAAI, NASNetLarge:zoph_2017}. Instead, o
Our main novelty is in the setup (shown in Fig. \ref{fig:example}) that provides images for crack detection with up to 5 lighting conditions with the help of illumination sources placed at multiple heights. The setup is portable and has been designed for field work with constrains in its maximum dimensions. To the best of our knowledge, there are no techniques that study the effect of height-varying illumination conditions for automatic visual inspection of defects. We do so by comparing the crack detection results for the different lighting conditions using customized as well as state-of-the-art classification architectures on the images of ceramic tiles acquired with our proposed setup. Here, we attempt to provide an insight about the illumination conditions that can help in improving future techniques for crack detection.

The paper is organized as follows. In Section \ref{sec:setup} we provide the details of the proposed acquisition setup. The experimental pipeline is then described in Section \ref{sec:pipeline} followed by the metrics in Section \ref{sec:metrics} that we use for comparing the results. Section \ref{sec:results} discusses the results while the conclusion and the future work are given in Section \ref{sec:conclusion}.% The future work is discussed in Section \ref{sec:future}.

\section{Setup}
\label{sec:setup}

%\begin{figure*}[tb]
%	\centering
%	\subfigure[Description]{\includegraphics[width=0.48\textwidth]{./images/mws_pav001_003_p001.pdf}}  
%    \subfigure[Description]{\includegraphics[width=0.48\textwidth]{./images/mws_pav001_003_p001_1.pdf}}
%	\includegraphics[scale=0.6]{Avio1/CompleteSystem.pdf}
%	\caption{Pipeline of the proposed systems. The system receives in input the acquired image. %The outputs are the sub-parts checked and the defect recognized.}
%	%\vspace{-0.5cm}
%	\label{fig:lightsframe}
%\end{figure*}
%Geometric constrains

In this section we describe the ad-hoc hardware development for this work. Our proposed setup consists of:
\begin{itemize}
    \item 3D printed illumination frame;
    \item addressable LED strips;
    \item a machine vision camera;% and 
    \item optically rectified lens.
\end{itemize}
The illumination frame, depicted in Fig. \ref{sfig:device_setup_lights}, hosts four lines of LED's strips which illuminate the object (i.e., a ceramic tile in our case) from all the four sides at different levels of height. With the press of a button, the setup acquires 5 images of the tile in the following lighting configurations:
\begin{itemize}
	\item all lights switched on,
	\item only level 1 (LED level closest to the tile) switched on,
	\item only level 2 (LED level above level 1) switched on,
	\item only level 3 (LED level above level 2) switched on, and
	\item only level 4 (LED level above level 3 and furthest from the tile) switched on.
\end{itemize}

Our design is inspired by the so-called \textit{dark field illumination} \cite{VanDommelen_1975} which tries to increase the contrast between the background and foreground regions using oblique lighting. The intuition behind this design is that cracks can be better visualized under specific lighting conditions based on the illumination angle. The four illumination levels provide different angles of illumination with a total of 189 white colored LEDs that can be activated individually or in pre-programmed multi-led illumination patterns. The goal is that of improving the visibility of defects on the tile and avoiding specularity.% The LEDs are SK9822 white LEDs integrated in a Adafruit DotStar Digital LED strip with 60 led/meter powered at 5 Volts.

The LEDs control is provided by an Arduino Nano board running a dedicated software routine. This routine runs in the Arduino microcontroller and executes the messages corresponding to the control actions given by the user. Here, a computing device connected through the USB serial connection is used for receiving the control actions from the user and sending the corresponding messages to the Arduino microcontroller for execution.
% The routine runs in the Arduino microcontroller (ATmega32u4) by executing messages provided by a host tablet computer connected through the USB serial connection. 

% This setup has been designed for field work with constrains in its maximum dimensions.
This setup has been designed keeping in mind the dimension-constrains for deployment in a real industrial scenario for the inspection of ceramic tiles. For this reason the main goal of the design has been that of creating a compact device that can be easily transported by a human operator. The main problem in such situation comes from the optics geometry. In particular, we need to capture a surface of 20x20 cm (the dimensions of the surface) from a short distance. In the presented system, this translates into lens with few millimeters of focal lengths entering in the realm of ultra-wide lens. We achieve the desired compactness with a combination of a machine vision camera and an optically rectified lens. This helps us to have an extreme diagonal field of view (${\approx135}$ degrees) while having a very short distance (just 6cm) between the tile and the lens tip.% (as shown in Fig. \ref{fig:cameradistance}). 

%We opted for a combination of a 5Mpx machine vision camera Basler Pulse puA2500-14um with a MT9P031 imaging sensor and a optically rectified Theia lens SY125M (the complete integrated inspection device is depicted in Fig. \ref{fig:frame3}). This combination provides an extreme diagonal field of view of ~135 degrees with a distance to the tile from the lens tip of just 6 cm (see \ref{fig:cameradistance})(this distance outdo the declared lens producer minimum distance of 10 cm).

%\begin{figure}
%\centering
%% %\begin{subfigure}[b]{.49\linewidth}
%% %\includegraphics[width=\linewidth]{./images/frame1.png}
%% %   \caption{View one}\label{fig:frame1}
%% %\end{subfigure}
%% \begin{subfigure}[b]{.49\linewidth}
%% \includegraphics[width=\linewidth]{./images/frame2_rev.png}
%%     \caption{CAD model of the frame}\label{fig:frame2}
%% \end{subfigure}
%% \begin{subfigure}[b]{.49\linewidth}
%% \includegraphics[width=\linewidth]{./images/device_setup_lights}
%%     \caption{Frame with LED strips \& lens}\label{fig:frame3}
%% \end{subfigure}
%\includegraphics[width=0.49\linewidth]{./images/device_setup_lights}
%\caption{Illumination frame with the multiple LED levels.}
%\label{fig:frame}
%\end{figure}

% \begin{figure}[th!]
% 	\centering
% 	\includegraphics[width=0.4\linewidth]{images/cameradistance2.eps}
% 	\caption{Distance from the lens tip to the tile.}
% 	%\vspace{-0.5cm}
% 	\label{fig:cameradistance}
% \end{figure}

\section{Experimental pipeline}
\label{sec:pipeline}
\begin{figure*}[th!]
	\centering
	\includegraphics[scale=0.9]{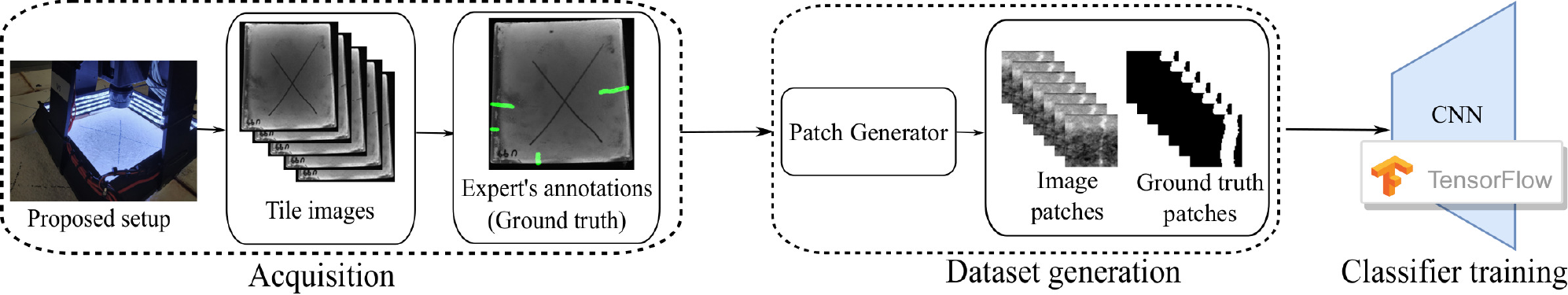}
	\caption{Experimental pipeline: First, the images with height-varying illumination are acquired using our setup. An expert then digitally annotates the crack locations in these images for training. The dataset is then generated by extracting the image-and-annotation patch pairs using the patch generator. The extracted patch pairs are used to train a patch based classifier to detect presence (or absence) of cracks. For evaluation, the inference is done on patches extracted from the acquired images.}
	%\vspace{-0.5cm}
	\label{fig:pipeline}
\end{figure*}

The pipeline used for our experimental procedure is shown in Fig. \ref{fig:pipeline} and the components are discussed below.

% \subsection{Acquisition and Registration}
\subsection{Acquisition}
\label{subsec:acquisition_and_registration}
%The images are acquired in a controlled illumination manner by placing the proposed setup over a tile kept on an acquisition table.
%With the press of a button, the setup then acquires 5 images of the tile in the following lighting configurations:
%\begin{itemize}
%    \item all lights switched on,
%	\item only level 1 (LED level closest to the tile) switched on,
%	\item only level 2 (LED level above level 1) switched on,
%	\item only level 3 (LED level above level 2) switched on, and
%	\item only level 4 (LED level above level 3 and furthest from the tile) switched on.
%\end{itemize}
The portable setup is placed over a tile kept on an acquisition table for the acquisition to be performed in a controlled illumination manner. It then acquires 5 images of the tile in various lighting configurations as discussed in Section \ref{sec:setup}. The setup is then moved and placed over the next tile and this process is repeated to acquire images of all the tiles.
For every tile, all the 5 images are acquired without moving the setup. Therefore, there is no relative motion and these are automatically registered with respect to each other. For these tiles, ground truth is provided by an expert (from our industrial partner) in the form of digital annotations (as shown in Fig. \ref{fig:pipeline}) which includes crack contours and locations of the tile corners. With the acquired images and their respective ground truth, we are now ready to generate the dataset (of patches) to be used for traning and evaluation. The following subsection explains the extraction of labeled patches from the acquired images and their organization to create the dataset.

\subsection{Dataset Generation}
\label{subsec:dataset_generation}
The registered images and the binary ground truth image containing the crack contours are used to extract labeled patches in a sliding window fashion. During extraction, the image patches are labeled as either positive (patches containing cracks), negative (patches that do not have cracks) or ambiguous based on the proportion of crack pixels in the corresponding ground truth patch. The proportion of crack pixels ${p}$ in a ground truth patch ${\psi}$ of size ${m \times n}$ is defined as:
\begin{equation}
    p = \frac{1}{m * n}\sum_{i=1}^{m} \sum_{j=1}^{n} I( \psi(i,j) > 0 ),
    \label{eqn:prop_crack_pixels}
\end{equation}
where ${I(\bullet)}$ is an indicator function such that ${I(True) = 1}$, ${I(False) = 0}$ and ${\psi(i,j)}$ is the value of pixel at location ${(i,j)}$ in the ground truth patch ${\psi}$. Using this definition of proportion of crack pixels ${p}$, patches in the registered images are labeled as follows:
\begin{itemize}
	\item Negative patches: ${p < 0.1}$,
	\item Ambiguous patches: ${0.1 \le p < 0.2}$,
	\item Positive patches: ${p \ge 0.2}$.
\end{itemize}
\changemarker{Intuitively, only those patches that contain no crack pixels (i.e., ${p = 0.0}$) should be labelled negative. Thus, setting ${p = 0.0}$ should suffice in labelling the patches as either negative or positive. However, in practice, the annotations are a few pixels wider than the actual crack-width, due to which some patches are incorrectly labelled as positive. To correctly label such patches as negative, we set a lower threshold to 0.1. Secondly, patches along the crack boundaries that contain few crack pixels have a similar appearance to that of the negative patches. Such ambiguous patches increase the number of false positives when labelled as positive. To avoid this, we set a higher threshold to 0.2.}
% \changemarker{The above thresholds are chosen heuristically based on patch size and observed proportion of annotated pixels in the crack and non-crack patches.}
Only the generated positive and negative patches are used for training and evaluation of the classifiers.

\subsubsection*{Data Balancing}
\label{subsubsec:data_balancing}
%\noindent\textbf{Data Balancing: }
Since every tile has only few crack patches, the number of negative examples is substantially larger than the number of positive examples. Thus, we get an imbalanced dataset by considering all extracted positive and negative patches. If an imbalanced dataset is used for training, the trained classifier can be biased towards the majority class, which in our case is the negative class.
\changemarker{To avoid this problem, balancing the training data with random under-sampling has been considered as one of the effective, easy and widely accepted methods \cite{Buda_NN_2018}. Alternate approaches that use cost sensitive loss functions \cite{Wang_IJCNN_2016}, \cite{Focal_loss:Lin_ICCV_2017} with imbalanced training data did not show a significant change in the results in our case, but notably increased the training time. We have therefore adopted the former method and artificially balanced the patches extracted from every image by random undersampling, i.e.,\ we consider all the positive patches and randomly select an equal number of negative patches for balancing the dataset.}
% To avoid this problem, we artificially balance the patches extracted from every image by random undersampling, i.e.,\ we consider all the positive patches and randomly select an equal number of negative patches for balancing the dataset.

\subsubsection*{k-Folds}
\label{subssubsec:folds}
%\noindent\textbf{k-Folds: }
Patches extracted from the acquired images are divided into 10 folds. Each fold has a fixed set of tiles for training and testing. This ensures that the patches used for testing in a particular fold are not used for training in the same fold. Considering data balancing and folds discussed above, we use the following data during the different phases, viz., train, validation and test.
%\noindent For fold ${ = F_K}$:
\begin{itemize}
	\item[] For fold ${ = F_K}$
	\item[--] Train${_K}$: \{Balanced Positives, Balanced Negatives\} from tiles selected for training in ${F_K}$,
	\item[--] Validation${_K}$: \{Balanced Positives, Balanced Negatives\} from tiles selected for testing in ${F_K}$,
	\item[--] Test${_K}$: \{Imbalanced Positives, Imbalanced Negatives\} from tiles selected for testing in ${F_K}$.
\end{itemize}

\subsubsection*{Spatial Resolution}
\label{subsubsec:spatial_resolution}
%\noindent\textbf{Spatial Resolution: }
The images acquired using the camera and the corresponding ground truth have a resolution of ${3840 \times 2748}$.
%The respective ground truth images also have the same resolution.
To have an insight about performance of the different classifiers on VGA-resolution (which is available even with inexpensive cameras), we experiment with low-resolution inputs. This is done by downsampling the acquired images with a factor of ${0.1667}$ both in height and width so that it has size closer to the VGA-resolution, i.e., ${640 \times 480}$. Here, the patches of size ${50 \times 50}$ are extracted with a stride of ${10}$ pixels.
We also experiment with the acquired images in their original size (high-resolution) without any downsampling by extracting patches of size ${299 \times 299}$ with a stride of ${60}$ pixels. Thus, the number of patches used in both low and high-resolution experiments is approximately the same.

% In our high-resolution experiments we use the acquired images in their original size without any downsampling. Instead, patches of size ${299 \times 299}$ are extracted with a stride of ${60}$ pixels. Thus, the number of patches used in both low and high-resolution experiments is approximately the same.

\subsection{Training}
\label{subsec:training}
The training of classifiers is done using various architectures. These include two custom architectures and six state-of-the-art classification architectures \cite{VGG16:Simonyan15, Xception:Chollet_2017_CVPR, ResNet50:He_2016_CVPR, DenseNet121:Haung_2017_CVPR, InceptionResNet-v2:Szegedy_2017_AAAI, NASNetLarge:zoph_2017}. For the custom architectures training is done from scratch. For all other architectures, the training is performed by fine-tuning where the pre-trained weights have been learnt on the imagenet classification dataset \cite{ImageNet_2009_CVPR}. %Following is a brief description of these architectures.

\subsubsection*{Training from scratch}
\label{subsubsec:training_from_scratch}
\begin{figure}
	\centering
	\begin{subfigure}[b]{.45\linewidth}
	    \includegraphics[width=\linewidth]{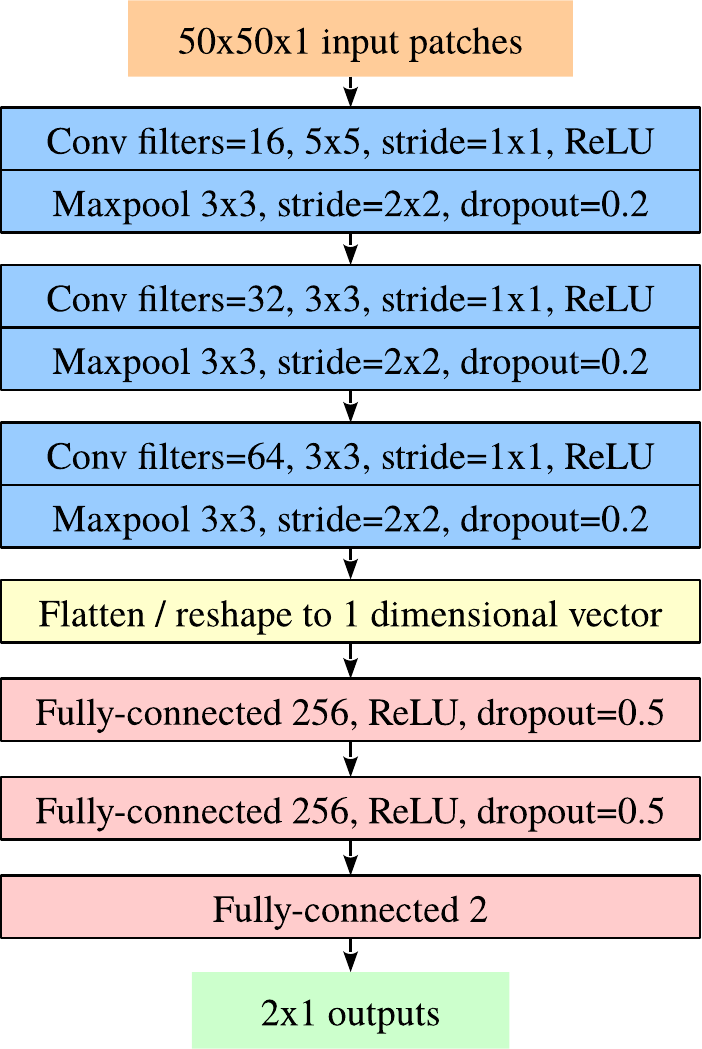}
	    \caption{TileNet6}\label{sfig:custom_arch1}
	\end{subfigure}
	\begin{subfigure}[b]{.45\linewidth}
	    \includegraphics[width=\linewidth]{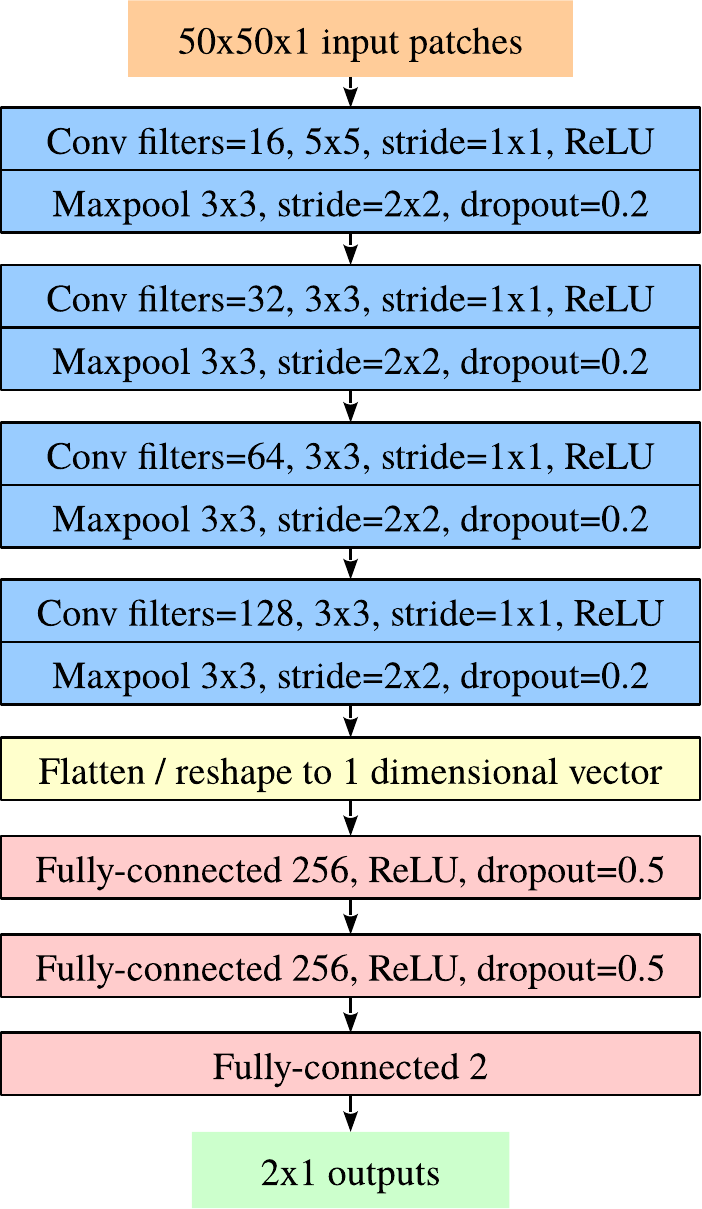}
	    \caption{TileNet7}\label{sfig:custom_arch2}
	\end{subfigure}
	\caption{Custom architectures considered for training.}
	\label{fig:custom_archs}
\end{figure}

%\noindent\textbf{Training from scratch: }
The two custom architectures that we train from scratch, i.e.,\ without using any pre-trained weights, have ${3-4}$ convolutional layers and three fully connected layers. The first of these architectures (\textit{TileNet6}) is shown in Fig. \ref{sfig:custom_arch1}. The second custom architecture only has one extra convolution layer in comparison to the first architecture. This was done to study the effect of having a slightly deeper architecture. We call this as \textit{TileNet7}, which is shown in Fig. \ref{sfig:custom_arch2}. Below, we also investigate the effect of having even more deeper architectures.
\begin{figure}
	\centering
	\includegraphics[width=\linewidth]{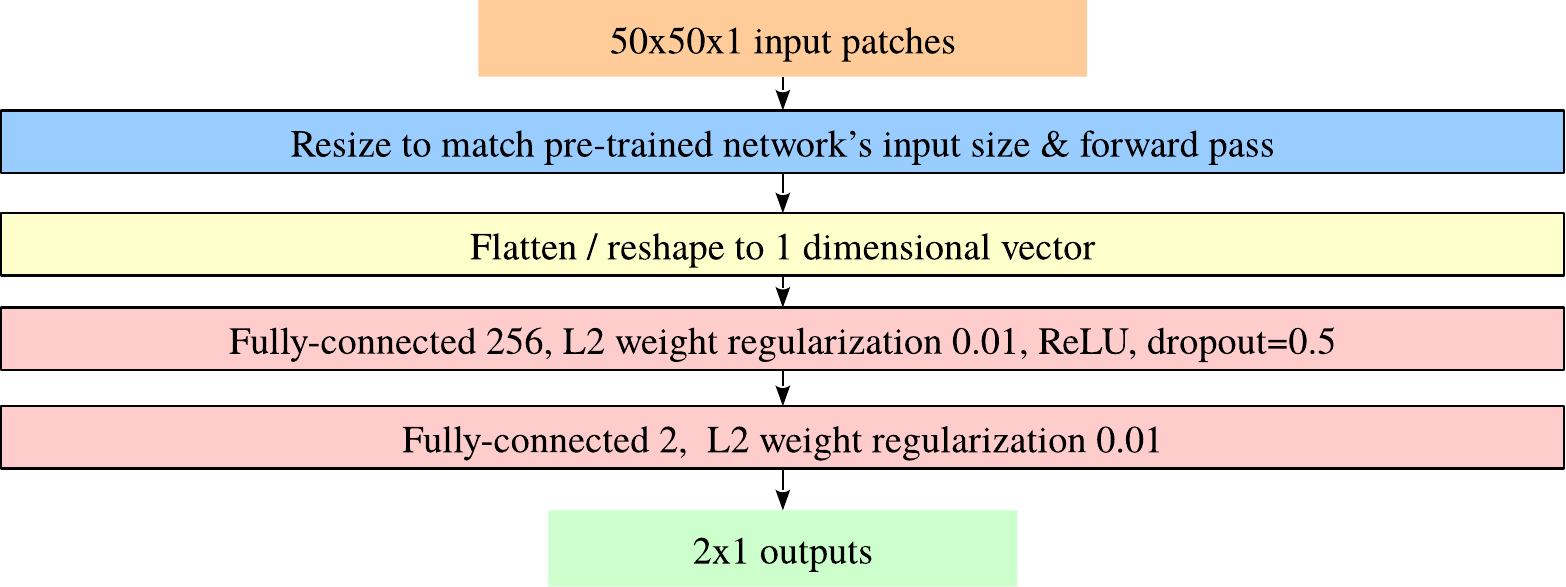}
	\caption{Architecture for fine-tuning.}
	\label{fig:transfer_learning}
\end{figure}

\subsubsection*{Fine-tuning}
\label{subsubsec:transfer_learning}
%\noindent\textbf{Fine-tuning: }
We have considered six state-of-the-art object classification architectures, viz. VGG16 \cite{VGG16:Simonyan15}, Xception \cite{Xception:Chollet_2017_CVPR}, ResNet50 \cite{ResNet50:He_2016_CVPR}, DenseNet121 \cite{DenseNet121:Haung_2017_CVPR}, InceptionResNet-V2 \cite{InceptionResNet-v2:Szegedy_2017_AAAI} and NASNetLarge \cite{NASNetLarge:zoph_2017}. These have outperformed most of the competing methods in the Imagenet Large Scale Visual Recognition Challenge (ILSVRC) \cite{ISLVRC:Russakovsky_2015_IJCV}. Their pre-trained weights (except for the top fully connected decision layer) have been downloaded from \textit{Keras Applications} \cite{Keras:chollet2015}.

First, we use these networks as feature extractors, i.e.,\ we feed them with the image patches and run a forward pass using the pre-trained weights. The extracted features and the corresponding labels are then used to train a shallow two-layered fully connected network which performs classification. This architecture is shown in Fig. \ref{fig:transfer_learning}.
Note that here the pre-trained weights are fixed and never updated. With the same network architecture we also try to fine-tune the pre-trained weights of certain layers. In this case we add a suffix ${finetune\_<layer\_name>\_onwards}$ to the model name. For example, when we perform transfer learning on ResNet50 by fine-tuning all the layers from ``conv\_5x'' onwards, the new model is named as ``ResNet50\_finetune\_conv\_5x\_onwards''. Likewise, if all the layers are fine-tuned, the suffix ${finetune\_all}$ is added to the model name. In addition, the models trained using high-resolution (i.e., without any downsampling) patches have the suffix ${\_HR}$. We use this nomenclature to discuss the results later in Section \ref{sec:results}. In the next section we describe the metrics used for performance evaluation.

\section{Evaluation Metrics}
\label{sec:metrics}

Since the training is performed on patches, it is logical to use patch-level metrics. However, since we eventually intend to evaluate the performance of classifiers to detect cracks in the whole tile, the use of image-level metrics is also proposed. The definitions of metrics used for evaluation at patch-level and image-level are as given below.

\subsection{Patch-level Metrics}
\label{ssec:patch_level_metrics}

%Considering the notations given below,
%\begin{itemize}
%	\item ${TP}$: Patches labeled as ``cracks'' and classified as ``cracks'' (i.e.,\ true positives)
%	\item ${FP}$: Patches labeled as ``not cracks'' but classified as ``cracks'' (i.e.,\ false positives)
%	\item ${TN}$: Patches labeled as ``not cracks'' and classified as ``not cracks'' (i.e.,\ true negatives)
%	\item ${FN}$: Patches labeled as ``cracks'' and classified as ``not cracks'' (i.e.,\ false negatives)
%\end{itemize}
%the patch-level metrics are discussed as follows.
Denoting true positives, false positives, true negatives and false negatives with ${TP}$, ${FP}$, ${TN}$ and ${FN}$, respectively, the patch-level metrics are discussed as follows.
\subsubsection{Accuracy}
\label{sssec:accuracy}
The first patch-level metric ${accuracy}$ is defined as follows:
\begin{equation}
	accuracy = \frac{TP + TN}{TP + FP + FN + TN}.
	\label{eq:accuracy}
\end{equation}

\subsubsection{Matthews Correlation Coefficient}
\label{sssec:mcc}
The accuracy metric can be misleading in the presence of imbalanced data. Therefore, we also use the \textit{Matthews Correlation Coefficient} (${MCC}$) \cite{mcc_metric:Matthews_1975} which is a more robust measure in the presence of imbalanced data. This measure is defined as follows:
\begin{equation}
MCC = \frac{TP*TN - FP*FN}{\sqrt{(TP+FP)(TP+FN)(TN+FP)(TN+FN)}}.
\label{eq:mcc}
\end{equation}

\subsection{Image-level Metrics}
\label{ssec:image_level_metrics}
Several patches can represent a single crack. Thus, low accuracy and MCC may not always indicate incorrect detection of cracks in the tile. Therefore, there is a need to quantify how well the classifiers work for the tiles and not just for patches. To address this issue, we propose the image-level metrics.
Let,
\begin{equation*}
	\begin{array}{rl}
		G: &\text{Set of cracks (closed contours or connected}\\
		   &\text{components) in ground truth,} \\
		D: &\text{Set of detected cracks (closed contours or} \\
		   &\text{connected components) in the given image,} \\
		n(X): &\text{Number of elements in set } X\text{,} \\
		&\changemarker{(\text{${X}$ represents ${D, G}$ or any other set of cracks})}\\
		I(\bullet): & \text{Indicator function, } I(True) = 1; I(False) = 0,
	\end{array}
\end{equation*}
then the image-level metrics are as below.

\subsubsection{Crack Presence Accuracy}
\label{sssec:crack_presence}
The first image-level metric is \textit{crack presence accuracy}, which indicates how well the tile is marked as having/not-having cracks. It gives a measure of how accurately can the underlying technique detect the defective tiles. For a given tile ${t}$ the crack presence metric ${(PM)_{t}}$ is defined as below.
\begin{equation}
	(PM)_t = I( I(n(G_t)>0) == I(n(D_t)>0) ).
	\label{eq:image_level_PM}
\end{equation}
For ${N}$ tiles, crack presence accuracy (CPA) is calculated as:
\begin{equation}
	CPA = average(PM) = \frac{1}{N}\sum_{t=1}^{N} (PM)_t.
	\label{eq:image_level_CPA}
\end{equation}
In the best case, the ${CPA = 1.0}$ indicating that all the tiles having at least one crack were correctly identified. On the other hand, ${CPA = 0.0}$ indicates that no tile was correctly identified for having presence/absence of the cracks.

\subsubsection{Crack Count F1 Score}
\label{sssec:crack_count_f1}
The next proposed image-level metric quantifies how close is the number of correctly detected cracks to the number of actual cracks in the ground truth. In other words, it indicates how accurately the underlying technique can detect the cracks. This metric (which we call as the crack count F1 score) is most relevant when the goal is to determine how many cracks can be correctly detected in the given tile.
\begin{figure}
    \centering
    \includegraphics[width=0.7\linewidth]{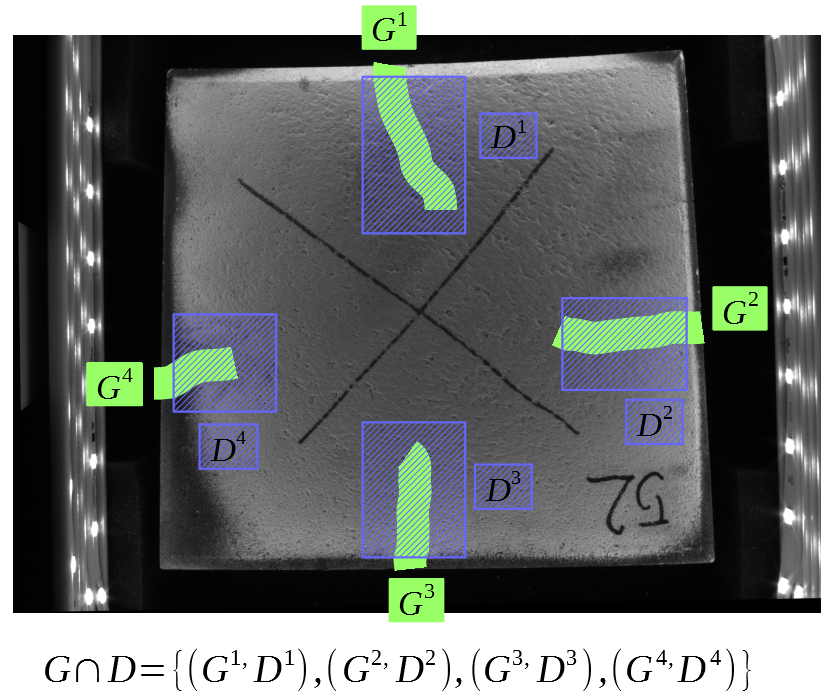}
    \caption{An example of associating correctly detected cracks. Ground truth cracks (${G^i}$) shown in green color are associated with overlapping detected cracks (${D^j}$) shown in violet color to get the set (${G \cap D}$) of correct detections.}
    \label{fig:associating_correct_detections}
\end{figure}

\begin{algorithm}
    \floatname{algorithm}{Procedure}
    % \caption{Steps to identify correctly detected cracks}
    \caption{Evaluation protocol for Crack Count F1 Score}
    \label{algo:correct_cracks}
    \begin{algorithmic}[1]
        \STATE Sort cracks in ground truth (${G}$) and detected cracks (${D}$) in descending order of the contour-area.
        \STATE For every ${i^{th}}$ crack in ${G}$ check overlap of every ${j^{th}}$ crack in ${D}$.
        \STATE Associate ${G^i}$ to ${D^j}$ for the largest overlap.
        \STATE ${G^i}$ remains unassociated if it has no overlap with any ${D^j}$.\hspace{-5pt}
        \STATE Once ${D^j}$ is associated to ${G^i}$, it is no more available for association with any other ${G^{k \ne i}}$.
        \STATE Find ${G \cap D}$ as set consisting of all the associated pairs ${\{(G^i, D^j)\}}$ which is nothing but the set of cracks correctly detected. One such example is shown in Fig. \ref{fig:associating_correct_detections}.
    \end{algorithmic}
\end{algorithm}
In order to calculate this metric, we first need to identify the cracks that are correctly detected, i.e., detected cracks that are also present in the ground truth. To do so we follow the procedure given in Procedure \ref{algo:correct_cracks}.
% \begin{itemize}
% 	\item Sort cracks in ground truth (${G}$) and detected cracks (${D}$) in descending order of the contour-area.
% 	\item For every ${i^{th}}$ crack in ${G}$ check overlap of every ${j^{th}}$ crack in ${D}$.
% 	\item Associate ${G^i}$ to ${D^j}$ for the largest overlap.
% 	\item ${G^i}$ remains unassociated if it has no overlap with any ${D^j}$.
% 	\item Once ${D^j}$ is associated to ${G^i}$, it is no more available for association with any other ${G^{k \ne i}}$.
% 	\item Find ${G \cap D}$ as set consisting of all the associated pairs ${\{(G^i, D^j)\}}$ which is nothing but the set of cracks correctly detected. One such example is shown in Fig. \ref{fig:associating_correct_detections}.
% \end{itemize}
Once the set of correctly detected cracks is available, we can calculate Recall, Precision and  F1 score for every tile ${t}$ as:
\begin{equation}
	\text{Recall }(R_t) =
	\left\{
	\begin{array}{ll}
	     \frac{n(G_t \cap D_t)}{n(G_t)}, & \text{ if } n(G_t) > 0,\\
	     1, & \text{otherwise,}%, i.e.,\ for zero cracks in ground truth,}
	\end{array}
	\right.
	\label{eq:image_level_Recall}
\end{equation}
\begin{equation}
	\text{Precision }(P_t) =
	\left\{
	\begin{array}{ll}
	     \frac{n(G_t \cap D_t)}{n(D_t)}, & \text{ if } n(D_t) > 0,\\
	     1, & \text{otherwise,}%, i.e.,\ for zero cracks in ground truth,}
	\end{array}
	\right.
	\label{eq:image_level_Precision}
\end{equation}
\begin{equation}
	F1_t = 2*\frac{P_t * R_t}{P_t + R_t}.
	\label{eq:image_level_F1}
\end{equation}
For ${N}$ tiles, the Crack Count F1 Score (CCF1) is calculated as the average of the F1 score given by:
\begin{equation}
	CCF1 = \frac{\sum_{t=1}^{N} F1_t * a_t}{\sum_{t=1}^{N} a_t}, \text{ where } a_t = n(G_t) + 1.
	\label{eq:image_level_CCF1}
\end{equation}
The maximum value ${CCF1 = 1.0}$ happens when the detected cracks are the same as the cracks present in the ground truth for all the tiles. Higher the value, more is the match between these quantities. In the worst case, i.e., when these quantities differ to their maximum we get ${CCF1 = 0.0}$. In this case none of the cracks are correctly detected in all the tiles.

\section{Results and Discussion}
\label{sec:results}

\begin{figure*}
	\centering
	\begin{subfigure}[b]{.49\linewidth}
	    \includegraphics[width=\linewidth]{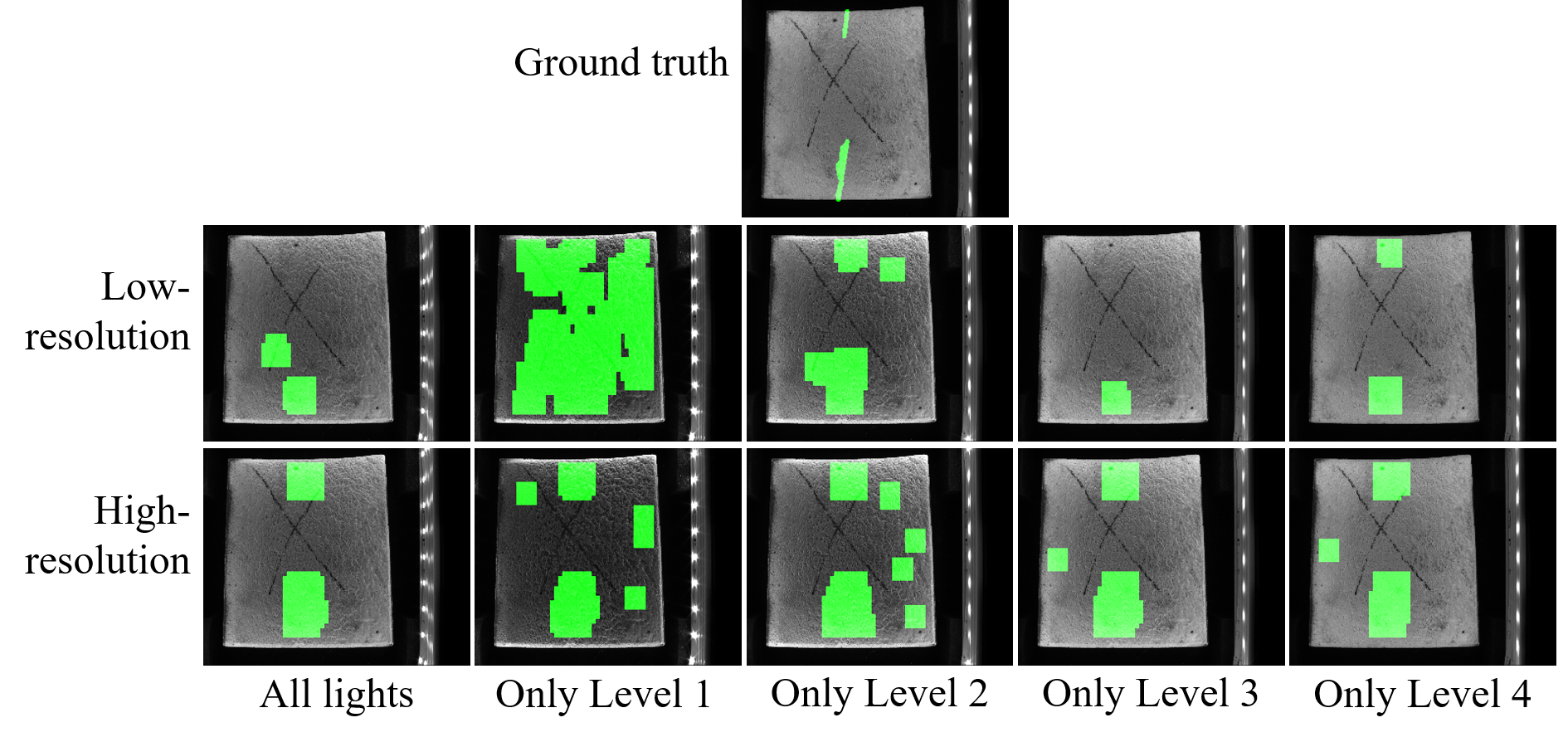}
	    \caption{}\label{sfig:visual_example1}
	\end{subfigure}
	\begin{subfigure}[b]{.49\linewidth}
	    \includegraphics[width=\linewidth]{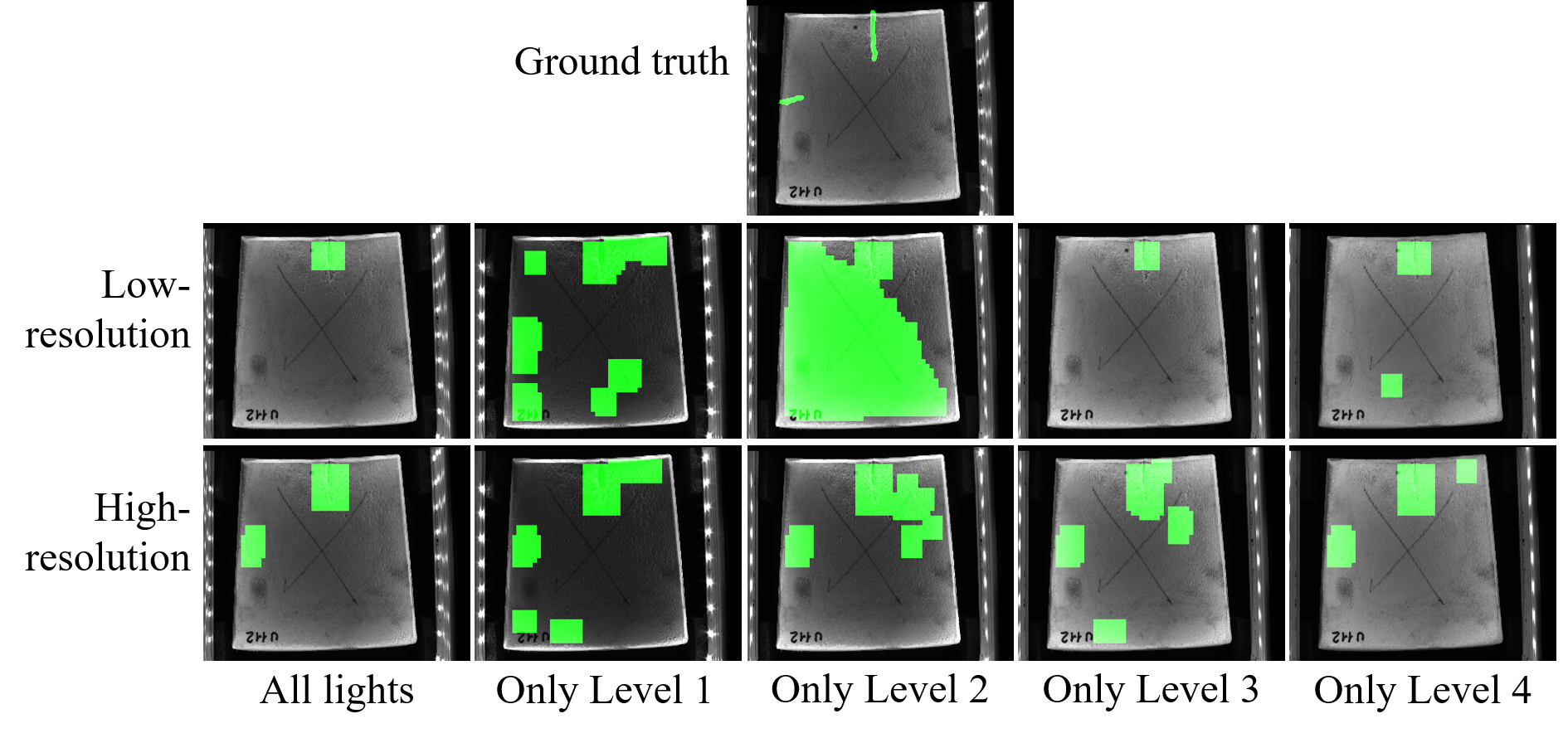}
	    \caption{}\label{sfig:visual_example2}
	\end{subfigure}
	\caption{Examples of crack detection for different illumination configurations for two tiles. The first row shows the ground truth, second row has results of classifiers trained using low-resolution patches while the third row has results of classifiers trained using high-resolution patches.}
	\label{fig:visual_examples}
\end{figure*}

\begin{figure*}
	\centering
	\includegraphics[width=\linewidth]{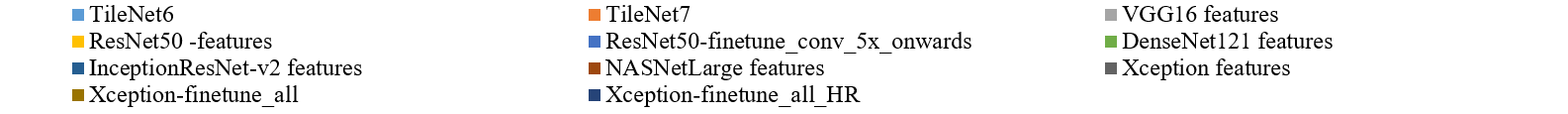}
	\begin{subfigure}[b]{\linewidth}
		\includegraphics[width=\linewidth]{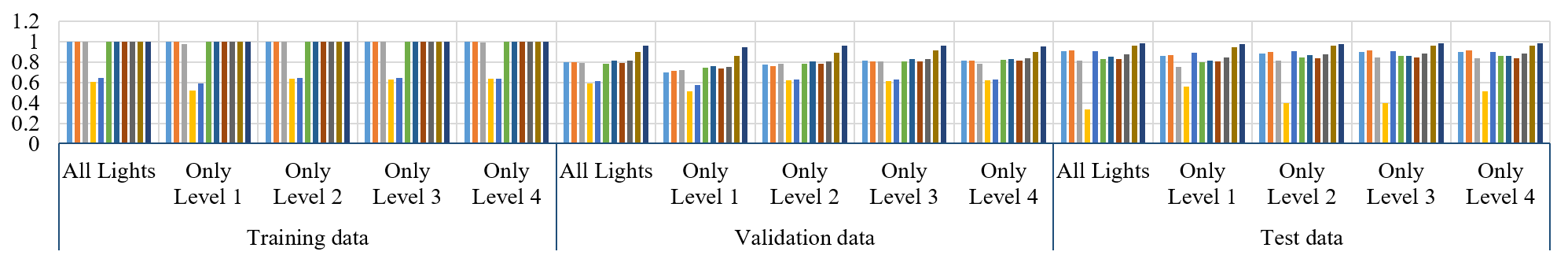}
		\caption{Average accuracy over 10 folds (0 to +1) for all models}\label{sfig:acc_all}
	\end{subfigure}
	\begin{subfigure}[b]{\linewidth}
		\includegraphics[width=\linewidth]{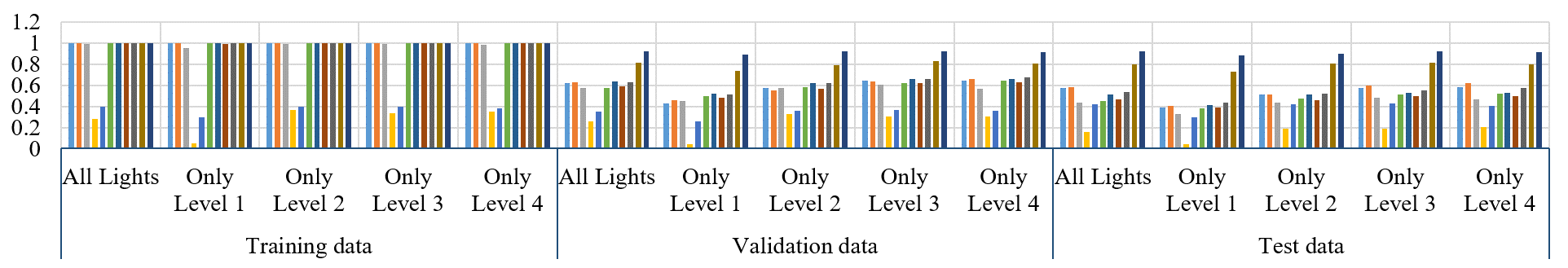}
		\caption{Average MCC over 10 folds (-1 to +1) for all models}\label{sfig:mcc_all}
	\end{subfigure}
	\begin{subfigure}[b]{\linewidth}
		\centering
		\includegraphics[width=0.7\linewidth]{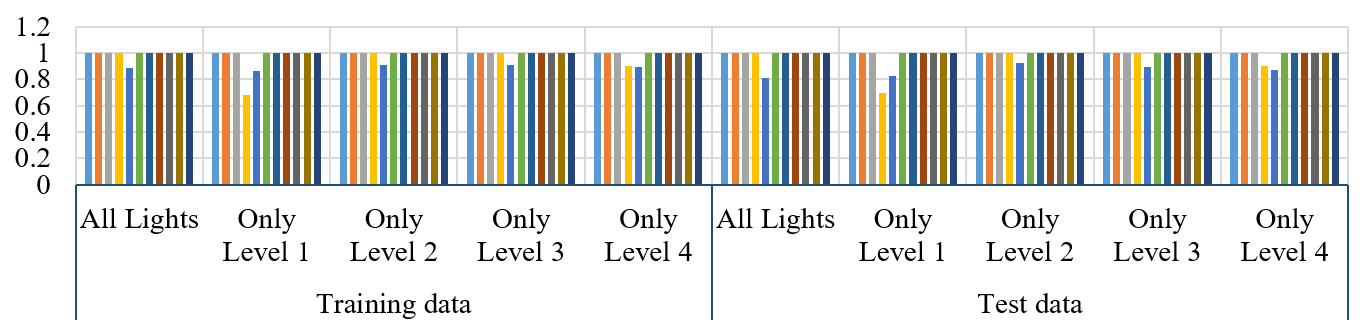}
		\caption{Average CPA over 10 folds (0 to +1) for all models}\label{sfig:cpa_all}
	\end{subfigure}
	\begin{subfigure}[b]{\linewidth}
		\centering
		\includegraphics[width=0.7\linewidth]{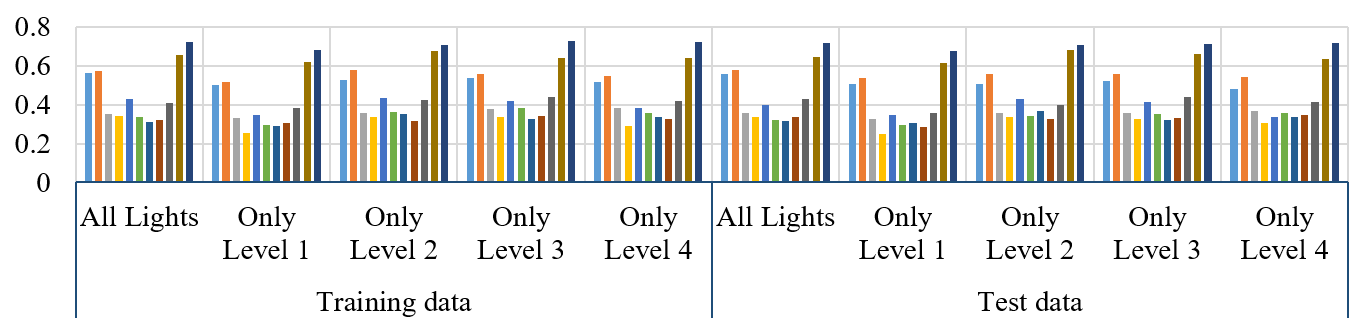}
		\caption{Average CCF1 score over 10 folds (0 to +1) for all models}\label{sfig:ccf1_all}
	\end{subfigure}
	\caption{Results for all the trained classifiers on low-and-high resolution patches using the different architectures and lighting configurations.}
	\label{fig:all_results}
\end{figure*}

Our dataset consists of 88 ceramic tile images acquired in five different lighting configurations using the procedure described in section \ref{subsec:acquisition_and_registration}. For each lighting configuration, the dataset is organized into 10 folds such that every fold is assigned about 70-76 and 8-12 tiles for training and test phases, respectively. Models corresponding to all the architectures discussed in section \ref{subsec:training} have been trained on a NVIDIA GeForce RTX 2080 Ti GPU using Adam optimizer \cite{adam_optimizer:Kingma_2015_ICLR} with a learning rate = 0.0001. The input size and batch size for training models using low-resolution patches is ${50 \times 50}$ and 128, respectively. For models trained using high-resolution patches, the input size is ${299 \times 299}$ while the batch size is reduced to 16 for managing the computational overhead. The training is done for about 1300 epochs in both cases where classifiers are trained using
\begin{inparaenum}[(a)]
    \item our custom architectures, and
    \item features extracted from pre-trained models.
\end{inparaenum}
The number of epochs is reduced to approximately 55 for the classifiers that are only fine-tuned over the pre-trained weights. The total time spent for training all the classifiers for the different lighting configurations, resolutions, folds and architectures is about 30 days. The results obtained by using these classifiers are presented below.

% All the above experiments were performed by using low-resolution inputs. To study the effect of using high-resolution inputs, we trained Xception\_finetune\_all\_HR by again fine-tuning every layer in the Xception architecture\footnote{We chose Xception architecture \cite{Xception:Chollet_2017_CVPR} in the high-resolution experiments because it provided the best results among all the transfer learning models trained using low-resolution inputs.}, but this time with the use of high-resolution inputs. For this experiment, the batch size was set to 16 for managing the computational overhead. Since only fine-tuning was required over the pre-trained weights (in both Xception\_finetune\_all\_LR and Xception\_finetune\_all\_HR), the number of epochs was reduced to approximately 55. In this case, the training time was about 2-3 days.

\subsection{Results}
\label{subsec:results_only}

Examples of crack detection on two tiles for the different lighting configurations are shown in Fig. \ref{fig:visual_examples}. The patch classification accuracy defined in Equation (\ref{eq:accuracy}) is compared for the trained models in Fig. \ref{sfig:acc_all}. Similarly, Fig. \ref{sfig:mcc_all} compares the Matthew's correlation coefficient defined in Equation (\ref{eq:mcc}) for these models. The crack presence accuracy defined in Equation (\ref{eq:image_level_CPA}) is compared in Fig. \ref{sfig:cpa_all}, while Fig. \ref{sfig:ccf1_all} compares the crack count F1 score defined in Equation (\ref{eq:image_level_CCF1}) for these models.

% Examples of crack detection on two tiles for the different lighting configurations are shown in Fig. \ref{fig:visual_examples}. The patch classification accuracy defined in Equation (\ref{eq:accuracy}) is compared for the trained models in Fig. \ref{sfig:acc_all}. Similarly, Fig. \ref{sfig:mcc_all} compares the Matthew's correlation coefficient defined in Equation (\ref{eq:mcc}) for these models. Fig. \ref{sfig:cpa_all}--\ref{sfig:ccf1_all} compare the image-level metrics. The crack presence accuracy defined in Equation (\ref{eq:image_level_CPA}) for training performed using low-resolution patches is shown in Fig. \ref{fig:CPA_LR}. A comparison of the same measure for models trained on the Xception architecture using high and low-resolution patches, respectively, is shown in Fig. \ref{fig:CPA_HR}. Similarly, Fig. \ref{fig:CCF1_LR}--\ref{fig:CCF1_HR} compare the crack count F1 score defined in Equation (\ref{eq:image_level_CCF1}) for these models.

\subsection{Discussion}
\label{subsec:discussion_only}

% We now discuss the results presented in the previous section for all the metrics defined in section \ref{sec:metrics}.

\subsubsection*{Accuracy}
\label{subsubsec:res_accuracy}

Looking at Fig. \ref{sfig:acc_all}, we observe that the training accuracy for almost all the models saturates to 1 while for validation it is about 0.8; an indication of overfitting. For test data also the accuracy is closer to 0.85 for most of the models. If only the feature-based models are considered, the model trained using Xception-features appears to be slightly better than the others. Nevertheless, the models trained using TileNet6 provides better accuracy in comparison to feature-based models. Also, having an additional convolution layer (i.e., TileNet7) helped in slightly improving the accuracy. Since the Xception-features have the best accuracy among the feature-based models, we also trained classifiers by fine-tuning all layers of the Xception architecture. These classifiers provided the best results reaching an average accuracy of 0.89 and 0.95 on the validation and test data, respectively. To study the effect of using high-resolution inputs, we also trained classifiers again by fine-tuning all layers of the Xception architecture but using high-resolution patches (Xception-finetune\_all\_HR). Here, we observe that the use of high-resolution inputs provides further improvement in the results. The average validation accuracy jumped from 0.89 to 0.95 while the average test accuracy improved from 0.95 to 0.98 and the training accuracy is also slightly improved.

% Yet, the best results were achieved by fine-tuning all layers of the Xception architecture resulting in an average accuracy of 0.89 and 0.95 on the validation and test data, respectively.

%, while ResNet50-features lead to poor accuracy. For ResNet50, fine-tuning more layers (ResNet50\_finetune\_conv5x\_onwards) slightly improves the accuracy, however, it is still not comparable with the other models\footnote{For ResNet50, the training and validation history showed high accuracy, but testing on the same data using the trained model resulted in poor accuracies. On the other hand, the accuracy for test data showed great improvement. Enabling batch normalization during the testing phase reversed this scenario, i.e., training and validation accuracies were high and matched with the history but test accuracy was very low. This seems to be a known batch normalization problem for ResNet50 in Keras (see \url{https://github.com/keras-team/keras/issues/6977}). Also, the preprocessing had to be disabled (see \url{https://github.com/fchollet/deep-learning-models/issues/96}).}.

% From the comparison shown in Fig. \ref{fig:accuracy_HR}, we observe that the use of high-resolution inputs improved the training accuracy by a small amount in most of the cases. Nevertheless, the average validation accuracy jumped from 0.89 to 0.95 while the average test accuracy improved from 0.95 to 0.98.

\subsubsection*{MCC}
\label{subsubsec:res_mcc}

%The MCC measure for the training, validation and test data for low-resolution experiments is shown in Fig. \ref{fig:mcc_LR}. Here, 

We observe that MCC (in Fig. \ref{sfig:mcc_all}) for most of the feature-based models is close to 0.5 (for both validation and test data), while it is slightly higher for the models trained on our custom architectures. The MCC for Xception architecture is comparable with that of our custom architectures. However, the MCC is highest for Xception\_finetune\_all with values 0.80 and 0.79 for for validation and test data, respectively. For the same, training with high-resolution patches shows further improvement with the MCC increasing from 0.80 to 0.91 for validation and from 0.79 to 0.90 for test data.

% Note that it is comparable to the MCC obtained using thermal images (where the highest reported MCC ${\approx}$ 0.80). When using high resolution patches for fine-tuning the Xception architecture (as shown in Fig. \ref{fig:mcc_HR}), the MCC increases from 0.80 to 0.91 for validation and from 0.79 to 0.90 for test. This indicates that the trained model is highly generalized and robust. Thus, fine-tuning the entire Xception architecture using high-resolution patches for training provides the best result in terms of MCC.

\subsubsection*{Crack Presence Accuracy}
\label{subsubsec:res_CPA}

Since the crack presence is quantified at image-level (as opposed to patch-level on which training is performed), the CPA is calculated for training data and test data that consists of images (and not image-patches) of tiles selected for training and test, respectively. Fig. \ref{sfig:cpa_all} shows that crack presence accuracy = 1.0 for for almost all the trained models in both training and testing phases. Similar is the case when using models trained on high-resolution patches. This metric can be more meaningful if 
\begin{inparaenum}[(a)]
	\item additional data containing tiles not having cracks is also considered for testing or
	\item the detection of cracks is to be performed in only specific regions of the tiles.
\end{inparaenum}
However, currently, we only have with us a dataset of tiles having cracks for which we perform crack detection over the complete tile-surface.

\subsubsection*{Crack Count F1 score}
\label{subsubsec:res_CCF1}

The crack count F1 scores shown in Fig. \ref{sfig:ccf1_all} indicate that the feature-based models are substantially less accurate in correctly detecting the cracks in comparison to those trained on TileNet6 \changemarker{(${CCF1 \approx 0.51}$)}. Adding an extra convolution layer (i.e., TileNet7) also helps in increasing this score \changemarker{(${CCF1 \approx 0.56}$)}. However, fine-tuning the entire Xception architecture leads to even better results \changemarker{(${CCF1 \approx 0.65}$)}. Also, the use of high-resolution inputs further increases the score \changemarker{(${CCF1 \approx 0.71}$)}, indicating a more accurate crack detection.

\subsubsection*{Effect of illumination sources placed at various heights}
\label{subsubec:effect_lighting_config}

The overall results in Fig. \ref{fig:all_results} indicate that both patch-level and image-level metrics improve as we move from \textit{Only Level 1} to \textit{Only Level 4}. In other words, higher the placement of the illumination source, better is the performance and less is the number of false positives (as seen in Fig. \ref{fig:visual_examples}). This trend is visible in all the results but more prominent for the models trained using high-resolution patches.
%This could be because certain cracks may not have clear visibility when the light source is placed closer to the tile's surface.
Secondly, the performance for \textit{Only Level 4} is more-or-less similar to that of \textit{All Lights}. This indicates that an illumination source placed at a greater height is as good as having a denser illumination setup (with sources at placed at various heights), for the purpose of crack detection.

\section{Conclusion and Future Work}
\label{sec:conclusion}

Our proposed height-varying illumination setup, which is designed for field work with constraints in its maximum dimensions, has been effectively used to acquire images of defective tiles. Crack detection has been performed on these images by means of patch-classification.
\changemarker{Our unique study on height-varying illumination conditions for crack detection suggests that lights placed at greater heights are more effective as compared to those placed near the tile's surface for crack detection.} In fact, their performance is as good as that of using together the lights at all the levels. \changemarker{Our study also indicates that fine-tuning of all the pre-trained weights of the Xception architecture provide the best results in comparison to all the other trained models across all the illumination condition. Moreover, use of high-resolution patches (i.e., without downsampling the acquired images) for training further improves the results. Thus, the intuition of performance improvement with the use of high-resolution patches is also validated across all the lighting conditions in our study. This should help in deciding the resolution versus performance trade-off when designing a real-time crack detection system for field use. The effectiveness of different illumination conditions on crack detection has been demonstrated using evaluation performed on classifiers trained with the state-of-the-art as well as our customized architectures.}

% Experiments have been carried out for binary classification to detect the presence of cracks in optical tile images. The performance in terms of all the metrics indicates that training models from scratch using our custom architectures is slightly better than using features extracted from popular pre-trained architectures. Moreover, the performance improves by increasing the model's depth. Nevertheless, for experiments considering low-resolution inputs with fine-tuning all the layers of Xception architecture provided the best results. Using inputs from the acquired images without downsampling them show further improvement in the classifier's performance. The classifiers trained using high-resolution inputs for fine-tuning all the layers of Xception architecture outperform all the other classifiers studied in this report. The corresponding models generalize well even without data augmentation. Another concluding remark is that use of only the lighting configuration wherein only light 1 is switched on, should be avoided as the performance of all the classifiers decreases for this configuration.

% \section{Future work}
% \label{sec:future}

The present work describes the experiments and results obtained from a relatively small number of tiles, which are difficult to procure. Also, the training images were annotated by highly specialized personnel in their very limited available time. Nevertheless, the effectiveness of portable setup has been clearly demonstrated by our experiments. In fact, its use can be generalized for automatic visual inspection of any object having a relatively planar surface.

% These tiles, obtained from a highly specialized industrial environment, are difficult to procure. Moreover, the training images must be annotated by highly specialized personnel.

% with a high monetary cost and a scarcity of available time. Nevertheless, in conjunction with our industrial partner, we are planing -and actually starting to execute- the expansion of the tiles samples by one order of magnitude. To do so, we are investing heavily in automating the procedure for the acquisition and annotation of the enlarged dataset.

Following encouraging preliminary results, in the future we will be focusing on using different sensor modalities towards extending the acquisition of the tiles by means of longwave infrared (LWIR) thermal sensors. Last but not least, we want to explore further the effect of both image and light resolution in the performance of the classification models.

%\section*{Acknowledgment}




\bibliographystyle{IEEEtran}
% \bibliography{IEEEabrv,references}
\bibliography{IEEEabrv,references_squeezed}


% \end{thebibliography}

\end{document}